\documentclass[runningheads]{llncs}

\usepackage{eccv}

\usepackage{eccvabbrv}

\usepackage{graphicx}
\usepackage{booktabs}
\usepackage{threeparttable}
\usepackage{multirow} 
\usepackage{longtable}
\usepackage{diagbox}
\usepackage{multicol}
\usepackage{makecell}
\usepackage{amssymb}
\usepackage{algorithm, algorithmic}
\usepackage{setspace}
\usepackage{graphicx}
\usepackage[accsupp]{axessibility}  % Improves PDF readability for those with disabilities.

\usepackage{hyperref}

\usepackage{orcidlink}
\usepackage{color}
\makeatletter
\def\thanks#1{\protected@xdef\@thanks{\@thanks
        \protect\footnotetext{#1}}}
\makeatother

\begin{document}

% ---------------------------------------------------------------
% TODO REVIEW: Replace with your title
\title{A Closer Look at GAN Priors: Exploiting Intermediate Features for Enhanced Model Inversion Attacks} 

% TODO REVIEW: If the paper title is too long for the running head, you can set
% an abbreviated paper title here. If not, comment out.
\titlerunning{A Closer Look at GAN Priors}

% TODO FINAL: Replace with your author list. 
% Include the authors' OCRID for the camera-ready version, if at all possible.
% \author{First Author\inst{1}\orcidlink{0000-1111-2222-3333} \and
% Second Author\inst{2,3}\orcidlink{1111-2222-3333-4444} \and
% Third Author\inst{3}\orcidlink{2222--3333-4444-5555}}
\author{Yixiang Qiu\inst{1, 2}$^{\dag}$\and
Hao Fang\inst{2}$^{\dag}$ \and
Hongyao Yu\inst{1}$^{\dag}$ \and
Bin Chen\inst{1,3,4}$^{\#}$  \and
Meikang Qiu\inst{5} \and
Shu-Tao Xia\inst{2,4}
\thanks{$^{\dag}$Equal contribution.}
\thanks{$^{\#}$Corresponding author.}
\thanks{This work was done while Yixiang Qiu was pre-admitted to Tsinghua University.}
}

% TODO FINAL: Replace with an abbreviated list of authors.
\authorrunning{Y. Qiu et al.}
% First names are abbreviated in the running head.
% If there are more than two authors, 'et al.' is used.

% TODO FINAL: Replace with your institution list.
% \institute{Princeton University, Princeton NJ 08544, USA \and
% Springer Heidelberg, Tiergartenstr.~17, 69121 Heidelberg, Germany
% \email{lncs@springer.com}\\
% \url{http://www.springer.com/gp/computer-science/lncs} \and
% ABC Institute, Rupert-Karls-University Heidelberg, Heidelberg, Germany\\
% \email{\{abc,lncs\}@uni-heidelberg.de}}
\institute{$^{1}$ Harbin Institute of Technology, Shenzhen \\
$^{2}$ Tsinghua Shenzhen International Graduate School, Tsinghua University \\ $^{3}$ Guangdong Provincial Key Laboratory of Novel Security Intelligence Technologies \\ $^{4}$ Pengcheng Laboratory\quad $^{5}$ Augusta University \\
\email{qiuyixiang@stu.hit.edu.cn}, \email{fang-h23@mails.tsinghua.edu.cn}\\
\email{yuhongyao@stu.hit.edu.cn}, \email{chenbin2021@hit.edu.cn}\\
\email{qiumeikang@yahoo.com}, \email{xiast@sz.tsinghua.edu.cn}
}

\maketitle

\newcommand{\todo}[1]{\textcolor{blue}{TODO: #1.}}
\begin{abstract}
  Model Inversion (MI) attacks aim to reconstruct privacy-sensitive training data from released models by utilizing output information, raising extensive concerns about the security of Deep Neural Networks (DNNs). Recent advances in generative adversarial networks (GANs) have contributed significantly to the improved performance of MI attacks due to their powerful ability to generate realistic images with high fidelity and appropriate semantics. However, previous MI attacks have solely disclosed private information in the latent space of GAN priors, limiting their semantic extraction and transferability across multiple target models and datasets. To address this challenge, we propose a novel method, \textbf{I}ntermediate \textbf{F}eatures enhanced \textbf{G}enerative \textbf{M}odel \textbf{I}nversion (IF-GMI), which disassembles the GAN structure and exploits features between intermediate blocks. This allows us to extend the optimization space from latent code to intermediate features with enhanced expressive capabilities. To prevent GAN priors from generating unrealistic images, we apply a ${l}_1$ ball constraint to the optimization process. Experiments on multiple benchmarks demonstrate that our method significantly outperforms previous approaches and achieves state-of-the-art results under various settings, especially in the out-of-distribution (OOD) scenario. Our code is available at: 
  \textcolor{blue}{\href{https://github.com/final-solution/IF-GMI}{https://github.com/final-solution/IF-GMI}}
  % \url{https://github.com/final-solution/IF-GMI}
\end{abstract}
\section{Introduction}
\label{sec:intro}
In recent years, Deep Neural Networks (DNNs) have experienced unprecedented development and achieved tremendous success in a wide range of applications, including face recognition \cite{he2016deep}, personalized recommendations \cite{wu2017session}, and audio recognition \cite{conneau2020unsupervised}. While DNNs bring us many practical benefits, concerns \cite{yu2024gi,fang2023gifd,fang2024one, chen2022adversarial} about privacy and security have also been raised and drawn great attention. Recent studies have demonstrated that there is a certain risk of privacy leakage for DNNs as an adversary could reveal private information from these pre-trained models. Various types of novel privacy attacks \cite{zeng2023narcissus,qiu2020adversarial,li2019reinforcement} have been proposed, such as $\textit{membership inference attack}$ \cite{shokri2017membership,hu2022membership} and $\textit{gradient inversion attack}$ \cite{fang2023gifd,yu2024gi}. Among the new attack methods, Model Inversion (MI) attack \cite{fang2024privacy} poses a greater threat due to its powerful capability in recovering the privacy-sensitive datasets that are collected and utilized for model training.

\cite{fredrikson2014privacy} proposes the first MI attack to reconstruct sensitive features of genomic data and demonstrate that linear regression models are vulnerable to such privacy attacks. Subsequent studies \cite{fredrikson2015model,song2017machine,yang2019neural} have extended MI attacks to more Machine Learning (ML) models, but are still limited to models with simple structure and low-dimensional data such as grayscale images. Recent advances in the MI attack field have overcome the challenges in image data recovery by applying Generative Adversarial Networks (GANs) \cite{goodfellow2020generative}, resulting in the extension to DNNs with more complex structure and high-dimensional data such as RGB images. \cite{zhang2020secret} first introduces the GANs to MI attack scenarios, serving as image priors. To better reveal privacy-sensitive information, \cite{zhang2020secret} and subsequent GAN-based methods \cite{chen2021knowledge,yuan2022secretgen,yuan2023pseudo,wang2021variational} train GANs with publicly available datasets that have structural similarity with target private datasets. Furthermore, \cite{struppek2022plug} propose to leverage the public pre-trained GAN models (\eg, StyleGAN \cite{karras2019style}) as GAN priors, which have a stronger ability to generate high-resolution images and do not require a time-consuming training process. 

\begin{figure}[tbp]
\centerline{
\includegraphics[width=1.05\columnwidth]{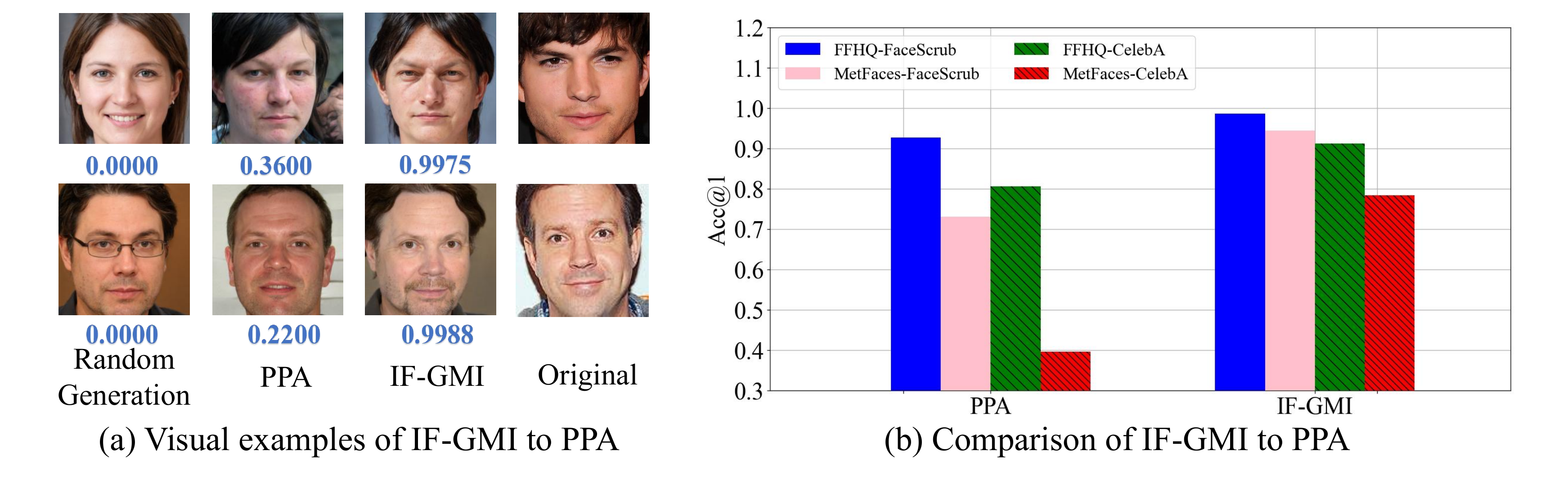}}
\caption{(a) Comparison of our proposed IF-GMI with baselines. The blue number below the images is the predicted confidence by the evaluation model. The first column shows the randomly generated images and the second column presents the reconstructed results by PPA \cite{struppek2022plug}, a typical GAN-based method focusing on directly optimizing the latent code of GAN model. The last two columns exhibit the results of our proposed IF-GMI and the ground truth images in the private dataset, respectively. (b) Top-$1$ attack accuracy of PPA and IF-GMI (ours) on four OOD scenarios.}
\label{intro}
\end{figure}

Although the aforementioned methods have achieved great progress in recovering high-quality and privacy-sensitive images, the effectiveness of GAN-based MI attacks is limited under certain scenarios. One typical challenge is the out-of-distribution (OOD) scenario, where there is a significant distributional shift between the target private dataset and the public dataset used in the training process of GAN priors. Most previous methods \cite{zhang2020secret,chen2021knowledge,wang2021variational,yuan2022secretgen} merely work well under scenarios with slight distributional shifts. For instance, they split the same dataset into two parts, one used as the public dataset and the other used as the private dataset. In recent years, some studies \cite{bau2018gan,tewari2020pie,yu2023editable,daras2021intermediate,shen2020interpreting} have demonstrated that there is rich semantic information encoded in the latent code and intermediate features of GANs. Inspired by these works, we empirically observe that the rich semantic information encoded in the intermediate features helps to sufficiently recover high-quality private data under more rigorous settings, as shown in Figure \ref{intro}. Therefore, it is imperative to explore methods for leveraging the GAN’s intrinsic layered knowledge into MI attacks, mitigating the OOD issue.

To this end, we propose a novel MI attack method,  \textbf{I}ntermediate \textbf{F}eatures enhanced \textbf{G}enerative \textbf{M}odel \textbf{I}nversion (IF-GMI), which effectively disassembles the GAN structure and leverages features between intermediate blocks. %In recent years, some studies \cite{bau2018gan,tewari2020pie,daras2021intermediate,shen2020interpreting} have demonstrated that there is rich semantic information encoded in the latent code and intermediate features of GANs. Inspired by these works, we extend the optimization objective and reformulate the MI attack from the perspective of  intermediate layer optimization problem. 
Specifically, we consider the generator of the GAN as a concatenation of multiple blocks and the vectors produced between the blocks as intermediate features. We first optimize the latent code input to the generator and then successively optimize the intermediate features from the start layer to the end layer.  To avoid unreal image generation, we utilize a ${l}_1$ ball constraint to restrict the deviation when optimizing the intermediate features. In the end, we collect the output images after each intermediate layer optimization process and select the final results with a simple strategy. 
%By exploiting the intermediate features, our method can utilize the rich semantic information encoded in GANs more sufficiently and recover high-quality private under more rigorous settings, as shown in Figure \todo{add visual examples}. 
We conduct comprehensive experiments to evaluate our method in multiple settings, including OOD scenarios, various target models, and different GAN priors. The encouraging experimental results demonstrate that the proposed method outperforms baselines on multiple metrics and achieves high attack accuracy on OOD settings. Finally, we perform extensive experiments and ablation studies to validate the effectiveness of the proposed method. Our main contributions are as follows:

\begin{itemize}
    \item We propose a novel GAN-based MI attack method, which disassembles the pre-trained generator and successively optimizes the latent code and intermediate features under the ${l}_1$ ball constraint.
    \item We demonstrate that our proposed method achieves state-of-the-art performance in a range of scenarios, especially under the challenging OOD settings.
    \item We conduct extensive experiments to validate the effectiveness and outstanding transferability of our method.
\end{itemize}

% (a) Comparison of $Acc@1$ metric under various settings of $L$ (\ie, the number of optimized layers). (b) Visual results generated from different end layers. We define $L=0$ as a special case that our method degenerates into merely optimizing the latent vectors $\mathbf{w}$.
\section{Related Work}
\subsection{GAN as prior knowledge}
GANs\cite{goodfellow2014generative} are a class of deep neural networks that consist of two functional components, a generator and a discriminator, trained concurrently through adversarial processes to generate realistic data. The objective of a GAN is to learn the distribution of the training dataset and generate more samples from the learned probability distribution\cite{goodfellow2020generative}. Well-trained GANs are able to generate high-fidelity and diverse images, excellent representative of which are StyleGANs \cite{karras2019style,stylegan2}. The generator of the StyleGAN consists of a mapping network  and a synthesis network. The former maps latent vectors into the intermediate latent space (\ie ${\mathcal{W}}$ space), and the latter generates images through style vectors. The feature in the ${\mathcal{W}}$ space is well-disentangled, which means that images sharing similar features correspond to analogous style vectors. Therefore, PPA\cite{struppek2022plug} performs their attacks by searching the style vectors in ${\mathcal{W}}$ space. 
% 
% For 中间层，估计可解释为什么中间的层比其它的层好
The style vectors in the front layers tend to control high-level aspects of the generated images like pose, face shape, and general hair style, while those in the back ones have more influence on details\cite{karras2019style}, such as smaller scale facial features and eyes open/closed.
% For OOD
Moreover, style vectors in ${\mathcal{W}}$ space do not need to follow the same distribution with the training data, which means that more diverse images can be generated by controlling the vectors\cite{karras2019style}. 

% Motivated by these properties, our investigation delves into the potential of leveraging intermediate latent space of different layers to augment MI attacks. Our findings reveal that this approach not only significantly improves attack accuracy but also ensures the preservation of high-quality inversion outcomes, particularly in the out-of-distribution scenario.

Recent works \cite{park2020generator,zhong2024hierarchical,fang2023gifd} have shown the richness of intermediate features in GANs, our investigation also tries to explore the potential of leveraging intermediate latent space of different layers to enhance MI attacks. Our findings reveal that this approach significantly improves attack accuracy and obtains high-quality inversion results, particularly under the harder OOD scenario.

% intermediate features of GANs

\subsection{Model Inversion Attacks}

Model inversion (MI) attacks aim at reconstructing the private training data from a trained model. Typically, MI attacks can be divided into the white-box scenario \cite{zhang2020secret} and black-box scenario \cite{kahla2022label}. 
We only focus on the white-box scenario in this paper, which means that the attacker has full access to the trained model.
% The earliest approach attempt to recover genomic markers from a linear regression model \cite{fredrikson2014privacy}.
This kind of attack is initially demonstrated through an attempt to extract genomic markers from a linear regression model, as highlighted in the earliest research by \cite{fredrikson2014privacy}.
% 
% Subsequent studies \cite{fredrikson2015model,song2017machine,yang2019neural} extend MI attacks on shallow networks and simple data (e.g. low-resolution greyscale image). However, as the complexity of data and models grows, the effectiveness of these algorithms for attacks decreases significantly. 
Building on this foundation, subsequent researches\cite{fredrikson2015model,song2017machine,yang2019neural}  have broadened the scope of MI attacks, applying them to more machine learning models like shallow networks, and simple forms of data, such as low-resolution grayscale images. However, as the scale of both the data and the models increases, the efficacy of MI attack methods diminishes dramatically.

% To attack deeper and wider DNNs and recover more complex data (e.g. RGB images), GMI\cite{zhang2020secret}, a GAN-based method, is proposed to train a GAN model with public data to capture image distributional prior information and generate images with high quality. The attackers first randomly sample a batch of latent vectors as the inputs of the GAN to generate dummy images, and feed them into the target image classifier to get the prediction. Then they optimize the input latent vectors by minimizing the cross-entropy loss between the prediction and the target class and the discriminator loss. 
% 
In response to this challenge, a novel approach known as GMI, introduced by \cite{zhang2020secret}, employs a GAN-based methodology to enhance the ability of MI attacks with deeper and wider DNNs. This innovative strategy leverages a GAN model trained on publicly available data to encapsulate the distributional characteristics of image data, thereby facilitating the generation of high-quality image reconstructions. The process involves the attackers first generating a set of preliminary images by inputting a batch of randomly sampled latent vectors into the GAN. These generated images are then fed into the target image classifier to obtain initial predictions. To refine the attack, the attackers iteratively optimize the input latent vectors. This optimization process aims to minimize the discrepancy between the classifier's predictions and the intended target class, as measured by the cross-entropy loss, while also reducing the discriminator loss. With the help of the GAN, GMI seeks to achieve more precise and convincing reconstructions of complex data, thereby representing a significant advancement in the field of MI attacks.

Lots of researches in recent years improve the attack performance on the white-box scenario based on GMI. SecretGen\cite{yuan2022secretgen} explores the scenario when the attackers know some auxiliary information about the private data.
KEDMI\cite{chen2021knowledge} improves the discriminator by incorporating target labels and recover the distribution of the input latent vectors for a target class.
VMI \cite{wang2021variational} reformulates the MI attack from the perspective of variational inference and introduce KL-divergence as a regularization to better approximate the target distribution with a variational distribution.
PPA\cite{struppek2022plug} employs pre-trained StyleGAN2 to reduce the time cost of attacks and extend the attacks to high-resolution images thanks to the excellent generative ability of StyleGAN2. Moreover, they propose a set of strategies to heighten attack accuracy and robustness, including initial selection, post-selection, and data augmentation. 
LOMMA\cite{nguyen2023re} introduces model augmentation into MI attacks to reduce overfitting of the target model. They train some surrogate models from the target model via model distillation, co-guiding the optimization process with improved loss function.
PLGMI\cite{yuan2023pseudo} proposes a top-$n$ selection strategy, using target models to generate pseudo labels for publicly available images, thereby directing the training process for the conditional GAN.

\begin{figure}[tbp]
\centerline{\includegraphics[width=\columnwidth]{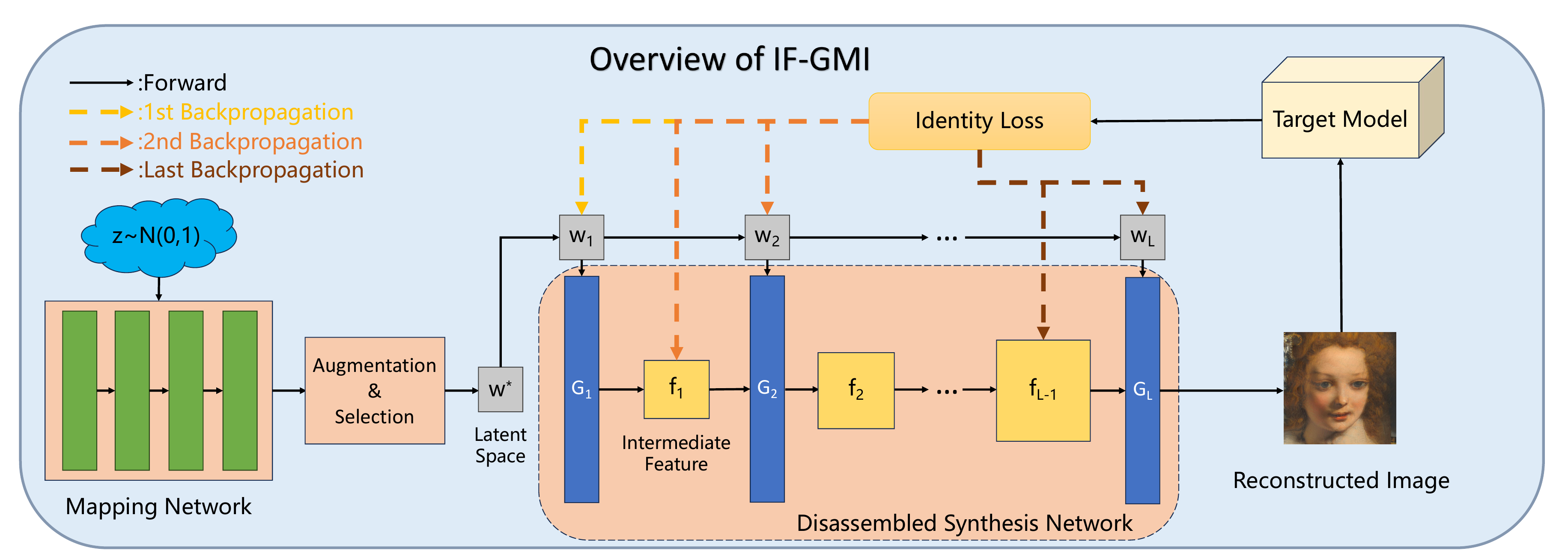}}
\caption{Overview of our proposed IF-GMI. Firstly, the latent vectors are sampled from standard Gaussian distribution and mapped into disentangled latent codes with semantic meanings by Mapping Network. Then we perform random augmentation on these latent codes to select optimal ones denoted as $\mathbf{w}^*$ for optimization. The Synthesis Network is disassembled into multiple blocks to search the intermediate features, which are successively updated with the identity loss calculated from the target model. Finally, the reconstructed images are generated from the last layer as results.}
\label{pipeline}
\end{figure}

\section{Methodology}
In this section, we begin by explaining the fundamental paradigm of MI attacks and provide a formulation for the MI problem. Subsequently, we present our main components and elaborate the detailed pipeline of the proposed IF-GMI, which contributes to the improved performance under the OOD scenario. See Figure \ref{pipeline} for an overview of our method.

\subsection{Preliminaries}
In this paper, we focus on the MI attacks under white-box settings, which means all the parameters and components of target models are available to the attacker. For image classification tasks, the malicious adversary aims to reconstruct privacy-sensitive images by leveraging the output prediction confidence of the target classifier and other auxiliary priors. Early works \cite{yin2020dreaming} directly optimize pixels in randomly sampled dummy images $\mathbf{x}$ to approximate target images $\mathbf{x}^{*}$ given the target model $T_{\theta}$ and target label $c$, which can be formulated as follows:
\begin{equation}
\begin{aligned}
    \mathbf{\hat{x}} = \mathop{\arg\min}\limits_{\mathbf{x}} \mathcal{L}(T_{\theta}(\mathbf{x}),c),
\end{aligned}
\end{equation}
where $\hat{\mathbf{x}}$ is the reconstructed image, $\mathcal{L}(\cdot,\cdot)$ denotes the classification loss designed for image optimization and $T_{\theta}(\mathbf{x})$ represent the output confidence. Due to the full access to the target model in white-box settings, the attacker can calculate loss and directly perform backpropagation to update dummy images.  

However, the methods above are no longer functional when $\mathbf{x}$ turns into high-dimensional data which has excessive search space. To tackle such issues, recent studies \cite{zhang2020secret,chen2021knowledge, struppek2022plug,yuan2023pseudo} introduce GANs as image priors due to their superior capability to generate high-fidelity RGB images. They propose to train a specially designed GAN with publicly available datasets that have structural similarities with the private dataset or utilize a public pre-trained GAN before the attack. Furthermore, the optimization objective is replaced with the latent vectors $\mathbf{z}$ of the generator, which has fewer parameters to optimize. With the aforementioned techniques, the MI problem is transformed into the following formulation:
\begin{equation}
\begin{aligned}
    \mathbf{\mathbf{\hat{z}}} = \mathop{\arg\min}\limits_{\mathbf{z}} \mathcal{L}_{id}(T_{\theta}(G(\mathbf{z}),c) + \lambda\mathcal{L}_{aux}(\mathbf{z}),
\label{loss}
\end{aligned}
\end{equation}
where $G$ represents the trained generator, $\mathcal{L}_{id}(\cdot,\cdot)$ denotes the identity loss calculated from the target model $T_{\theta}$ and $\mathcal{L}_{aux}(\cdot)$ is an optional auxiliary loss (\eg, the discriminator loss) with a hyperparameter $\lambda$. By minimizing the Eq.\ref{loss}, the adversary updates the latent vectors $\mathbf{z}$ into the optimal results $\mathbf{\hat{z}}$ and generate final images through $\mathbf{\hat{x}}=G(\mathbf{\hat{z}})$. 

Intuitively, directly optimizing the input latent code of GAN priors serves as a natural method to acquire ideal reconstructed images, leading to its widespread application in all the previous works. However, recent studies \cite{bau2018gan,tewari2020pie,daras2021intermediate,shen2020interpreting} have indicated that there is fairly rich semantic information in the intermediate features of GANs except for the input latent code. This inspires us to surpass the limitation of merely searching the latent space and propose a novel method focusing on the intermediate feature domains, which are more close to the output.

\subsection{Exploiting Intermediate Features for Enhanced MI Attacks}

In the following part, we delve into the internal structure of the GAN prior, attempting to explore the hierarchical layers for enhanced utilization of the rich semantics learned by the generator. Following the pipeline shown in Figure \ref{pipeline}, we will elucidate each component in detail.

\paragraph{\textbf{The GAN prior.}} Most previous GAN-based attacks \cite{zhang2020secret,chen2021knowledge,yuan2023pseudo,nguyen2023re} require training a specialized GAN with essential auxiliary dataset towards the specific target classifier. However, the prior knowledge of GANs trained under the above setting will be excessively aligned with the target model and the auxiliary dataset, leading to significant reduction in transferability and generalization. 

Therefore, our method relies on the pre-trained StyleGAN2 \cite{karras2020training} instead of training a GAN from scratch. The generator of StyleGAN2 can be simply divided into two components, consisting of a mapping network $G_{map}:\mathcal{Z}\xrightarrow{}\mathcal{W}$ which maps the initial latent vectors $\mathbf{z}\in{\mathcal{Z}}$ into the extended $\mathcal{W}$ space \cite{abdal2019image2stylegan}, and a synthesis network $G_{syn}:\mathcal{W}\xrightarrow{}\mathcal{X}$ which generates images $\mathbf{x}$ with mapped vectors $\mathbf{w}\in{\mathcal{W}}$. Due to the reduced feature entanglement in $\mathcal{W}$ space that facilitates better style generation, we set $\mathbf{w}$ as the initial optimization objective rather than the commonly used latent code $\mathbf{z}$ in previous works. Specifically, we first randomly sample a batch of latent vectors $\mathbf{z}$ from Gaussian distribution and then map them with $G_{map}$ to acquire $\mathbf{w}$, which will be iteratively updated in the first step of intermediate features optimization. Moreover, the StyleGAN2 is pre-trained without the utilization of the target model $T_{\theta}$ or other auxiliary prior corresponding to the target dataset, ensuring the flexibility and transferability of our method when attacking different target models and datasets.

\paragraph{\textbf{Initial Selection.}} Owing to the randomness in sampling latent vectors $\mathbf{z}$, it is potential part of them cannot facilitate the generation of appropriate images, leading to a decrease in attack accuracy. To reduce the risk of generating misleading and low-quality images, previous studies \cite{yuan2022secretgen,an2022mirror,struppek2022plug} have explored the technique of initial selection and validated its effectiveness in obtaining robust latent vectors. Specifically, we first generate images with the randomly samples $\mathbf{z}$, apply a series of transformations $Aug(\cdot)$ to the images, and feed them into the target classifier $T_{\theta}$ for corresponding prediction confidence. By selecting the latent vectors with higher scores, we can significantly improve the quality of the final images to better approximate the target distribution. 

Inspired by these prior studies \cite{yuan2022secretgen,an2022mirror,struppek2022plug}, we also include the initial selection technique in our method and apply standard image transformations, such as random cropping, resizing and flipping. Different from previous methods, we perform initial selection on the mapped vectors $\mathbf{w}$ instead of latent vectors $\mathbf{z}$. The robust vectors $\mathbf{w}$ are obtained with the following equation:
\begin{equation}
\begin{aligned}
    \mathbf{w}_{init} = \mathop{\arg\max}\limits_{\mathbf{w}} \mathrm{Conf}(T_{\theta}(Aug(G_{syn}(\mathbf{w}))),c),
\end{aligned}
\end{equation}
where $\mathrm{Conf}(\cdot,\cdot)$ measures the confidence score for augmented images $Aug(G_{syn}(\mathbf{w}))$ given the specific label $c$. 

\begin{algorithm}[!ht]
\setstretch{1.15}
\caption{Pseudocode of the core algorithm in our proposed IF-GMI}\label{pseu}
\begin{algorithmic}[1]
\renewcommand{\algorithmicrequire}{\textbf{Input:}}
\renewcommand{\algorithmicensure}{\textbf{Output:}}
\REQUIRE $G_{syn}$: a pre-trained generator; $L$: the number of intermediate features; $T_\theta$: the target classifier; $\mathcal{L}_{id}$: the identity loss; $r[1\dots{L}]$: the radius value of $l_1$ ball for each hierarchical features; $N$: the number of iterations;
\ENSURE Reconstructed images $\mathbf{x}^*$;
\STATE Acquire latent vectors $\mathbf{w}_{init}$ via initial selection process
\STATE $\mathbf{w}_{(0)} \xleftarrow{} \mathop{\arg\min}\limits_{\mathbf{w}} \mathcal{L}_{id}(G_{syn}(\mathbf{w}_{init}))$
\STATE Decompose the $G_{syn}$ into $G_{L+1}\circ{G}_{L}\circ\cdots{G}_{2}\circ{G}_{1}$
\STATE Obtain the first intermediate feature $\mathbf{f}_{(1)}^0=G_{1}(\mathbf{w}_{(0)})$
\STATE Set $\mathbf{w}_{(1)}^{0}=\mathbf{w}_{(0)}$
\FOR{$i\xleftarrow{}1 \ $to$ \ L$} 
    \STATE Set $G_{remain}=G_{L+1}\circ G_{L}\dots \circ G_{i+1}$
    \FOR{$j\xleftarrow{}1 \ $to$ \ N$}
        \STATE $loss = \mathcal{L}_{id}(G_{remain} (\mathbf{f}_{(i)}^{j-1},\mathbf{w}_{(i)}^{j-1}))$
        \STATE $\mathbf{f}_{(i)}^{j} \xleftarrow{} Adam(\mathbf{f}_{(i)}^{j-1};loss), ||\mathbf{f}_{(i)}^{j}-\mathbf{f}_{(i)}^0||_1 \leq r[i]$
        \STATE $\mathbf{w}_{(i)}^{j} \xleftarrow{} Adam(\mathbf{w}_{(i)}^{j-1};loss), ||\mathbf{w}_{(i)}^{j}-\mathbf{w}_{(i)}^0||_1 \leq r[i]$
    \ENDFOR
    \STATE $\mathbf{f}_{(i+1)}^{0} = G_{i+1}(\mathbf{f}_{(i)}^N,\mathbf{w}_{(i)}^N)$, $\mathbf{w}_{(i+1)}^{0} = \mathbf{w}_{(i)}^N$
\ENDFOR
\STATE The final images $\mathbf{x}^* = \mathbf{f}_{(L+1)}^{0}$ 
\RETURN $\mathbf{x}^*$
\end{algorithmic}
\end{algorithm}

\paragraph{\textbf{Intermediate Features Optimization.} }
According to the research of \cite{karras2019style}, the front blocks in the generator control the overall characteristics while the back ones have more influence on local details. Previous studies\cite{zhang2020secret,struppek2022plug,yuan2023pseudo} neglect the role of the latter, which limits the attack performance.
To take advantage of the individual blocks, we propose intermediate features optimization, as shown in the Algorithm \ref{pseu}.
% 
% Following the steps depicted in the Algorithm \ref{pseu}, 
We first optimize the selected latent vectors $\mathbf{w}_{init}$ to obtain the optimal ones $\mathbf{w}_{(0)}$ before launching intermediate features optimization. Then we disassemble the pre-trained generator into $L+1$ blocks for hierarchical layer searching, \ie, 
\begin{equation}
\begin{aligned}
    G_{syn}=G_{L+1}\circ{G}_{L}\circ\cdots{G}_{2}\circ{G}_{1}.
\end{aligned}
\end{equation}
And we can feed $\mathbf{w}_{(0)}$ into block $G_{1}$ to attain the first intermediate feature $\mathbf{f}_{(1)}^{0}$. 

For each intermediate block $G_{i+1},i\in[1,\dots,L]$, the corresponding intermediate features $\mathbf{f}_{(i+1)}^{0}$ are acquired with following steps. First, we generate images utilizing the remaining blocks ($\ie, \mathbf{x}_{i}={G}_{L+1}\circ{G}_{L}\dots{G}_{i+1}(\mathbf{f}_{(i)},\mathbf{w}_{(i)})$) and input them into the target classifier $T_{\theta}$ to compute the prediction confidence for loss function. Then, we repeat the aforementioned process to iteratively update both $\mathbf{w}_{(i)}$ and $\mathbf{f}_{(i)}$. During the optimization process, we restrict the $\mathbf{f}_{(i)}$ within the $l_1$ ball with radius  ${r}[{i}]$ centered at the initial intermediate feature $\mathbf{f}_{(i)}^{0}$ to avoid excessive shift that may lead to collapse image generation. Once the iteration process is completed, the optimized $\mathbf{w}^{N}_{(i)}$ and $\mathbf{f}^{N}_{(i)}$ are fed into the block $G_{i}$ to obtain the next intermediate features $\mathbf{f}_{(i+1)}^{0}$. Moreover, we denote the optimized $\mathbf{w}^{N}_{(i)}$ as the initial latent vector $\mathbf{w}_{(i+1)}^{0}$ before the next layer optimization starts.

Once we finish searching the last intermediate layer, we can generate the final images $\textbf{x}^*$ from the last intermediate feature $\textbf{f}_{(L)}^N$, \ie, $\textbf{x}^*=\textbf{f}_{L+1}^0=G_{i+1}(\textbf{f}_{(L)}^N)$.

\paragraph{\textbf{The Overall Loss.}} While the cross-entropy loss $\mathcal{L}_{CE}$ serves as the identity loss in most early works \cite{zhang2020secret,chen2021knowledge,yuan2022secretgen}, there is a major drawback of $\mathcal{L}_{CE}$. Specifically, the gradient vanishing problem emerges when the prediction confidence of target label $c$ approaches the ground truth in the one-hot vector. Following the previous study \cite{struppek2022plug}, we rely on the Poincar\'e loss function to overcome this problem. Therefore, the identity loss function utilized in out method is defined as follows:
\begin{equation}
\begin{aligned}
    \mathcal{L}_{id}=\mathrm{arccosh}\left(1+\frac{2||v_1-v_2||_{2}^{2}}{(1-||v_1||^{2}_{2})(1-||v_2||^{2}_{2})}\right),
\end{aligned}
\end{equation}
where $||v||_2$ is the Euclidean norm for the given vector. In our experiments, we denote $v_1$ as the normalized prediction confidence and $v_2$ as the one-hot vector for ground truth. Notably, the original number 1 in $v_2$ is substituted with 0.9999 to avoid division by zero.
\section{Experiments}

In this section, we first illustrate the details of our experimental settings. Then, we compare our method with state-of-the-art baselines to evaluate the attack performance. Furthermore, we conduct extensive experiments on multiple target datasets and models to further validate the effectiveness of our method in various settings. Finally, the ablation study will be evaluated on the first $100$ classes of the whole dataset due to cost concerns.

\subsection{Experimental Setup}

\paragraph{\textbf{Datasets.}} 
We evaluate our method on two classification tasks, including facial image classification and dog breed classification. For the facial image classification task, we select the FaceScrub\cite{facescrub} and CelebFaces Attributes\cite{celeba} (CelebA) as private datasets to train the target models. FaceScrub consists of facial images of actors and actresses with $530$ identities in total. CelebA contains facial images of $10177$ identities with coarse alignment. For FaceScrub, we utilize all the identities in the major experiment. For CelebA, we select the top $1000$ identities with the most images for our experiment, consisting of over 30000 images.
We use Flickr-Faces-HQ\cite{karras2019style} (FFHQ) and MetFaces\cite{karras2020training} as public datasets. FFHQ consists of $70000$ high-quality human face images. MetFaces is an image dataset of $1336$ human faces extracted from the Metropolitan Museum of Art Collection, which has a huge distributional shift with real human faces.
For the dog breed classification task, we use Stanford Dogs\cite{standforddog} as a private dataset and Animal Faces-HQ Dogs\cite{afhq} (AFHQ) as a public dataset. 
To adapt to the target model, all images in the various datasets are pre-processed to a resolution size of $224\times 224$ pixels in our experiment.

\paragraph{\textbf{Models.}}
% 
% We train various architectures of classification model separately on private datasets mention above, including Resnet-18\cite{resnet}, DenseNet-169\cite{densenet}, Resnet-152\cite{resnet} and ResNeSt-101\cite{zhang2022resnest} as target models and Inception-v3\cite{inceptionv3} as evaluation models. For generator, we consider StyleGAN2 with the prior public datasets respectively.

We trained a variety of classifiers on the private datasets mentioned above, including various architectures such as ResNet-18\cite{resnet}, DenseNet-169\cite{densenet}, ResNet-152\cite{resnet}, and ResNeSt-101\cite{zhang2022resnest}, as target models. Following the settings in the previous work \cite{struppek2022plug}, we select Inception-v3 \cite{szegedy2016rethinking} as the evaluation model. For the generative model, we employ publicly released StyleGAN2 pre-trained on the aforementioned public datasets.

\paragraph{\textbf{Metrics.}}

Following PPA\cite{struppek2022plug}, we evaluate the performance of our attack method on various kinds of metrics as follows:
\begin{itemize}
    \item \textbf{Attack Accuracy.} This metric serves as a criterion on how well the generated samples resemble the target class. We use the evaluation model trained on the same dataset with the target model to predict the labels on reconstructed samples and compute the top-$1$ and top-$5$ accuracy for target classes, denoted as $Acc@1$ and $Acc@5$ respectively. The higher the reconstructed samples achieve attack accuracy on the evaluation model, the more private information in the dataset can be considered to be exposed\cite{zhang2020secret}.
    \item \textbf{Feature Distance.} The feature is defined as the output of the model's penultimate layer. We compute the shortest feature $l_2$ distance between reconstructed samples and private training data for each class and calculate the average distance. The evaluated feature distances on the evaluation model and a pre-trained FaceNet\cite{schroff2015facenet} are denoted as $\delta_{eval}$ and $\delta_{face}$, respectively. 
    \item \textbf{Fr\'echet Inception Distance (FID).} FID\cite{fid} is commonly used to evaluate the generated images of GANs. It computes the distance between the feature vectors from target private data and reconstructed samples. The feature vectors are extracted by Inception-v3 pre-trained on ImageNet. The lower FID score shows higher realism and overall diversity \cite{wang2021variational}.
    \item \textbf{Sample Diversity.} We compute Precision-Recall\cite{pr} and Density-Coverage\cite{dc} scores, whose higher values indicate greater intra-class diversity of the reconstructed samples. Our results for these four
    metrics are stated and analyzed in the Appendix \ref{app:prdc}.
\end{itemize}

\begin{figure}[tbp]
\centerline{
\includegraphics[width=\columnwidth]{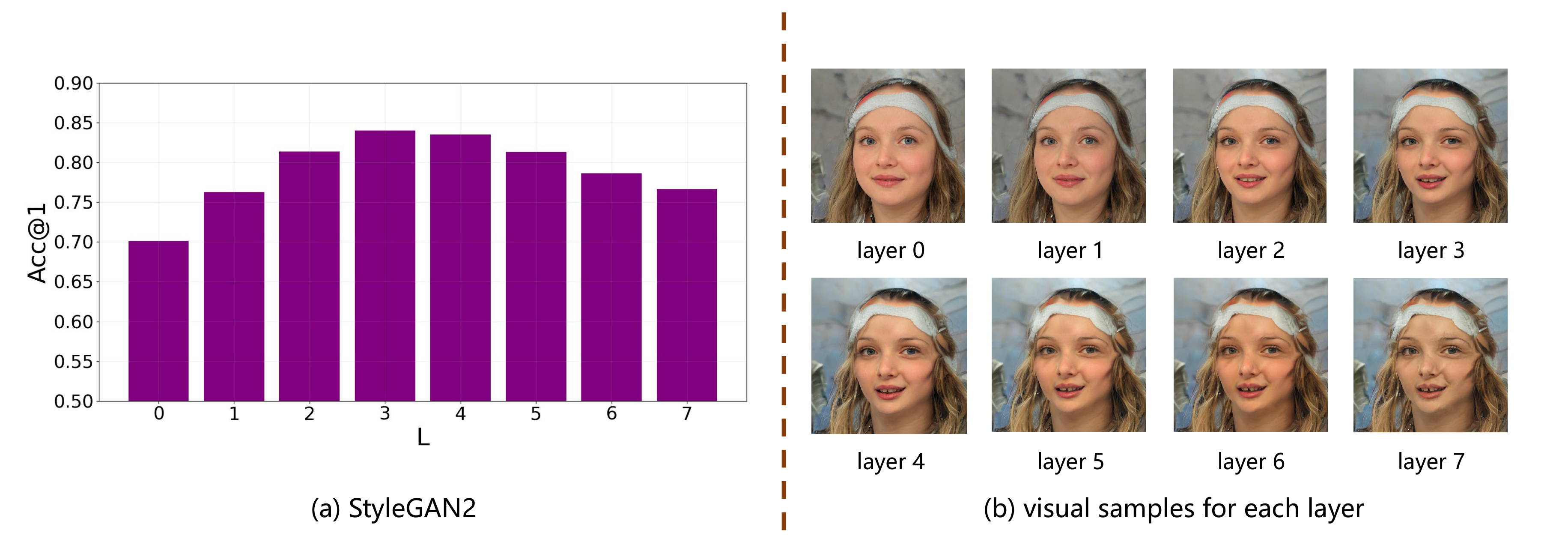}}
\caption{(a) Comparison of $Acc@1$ metric under various settings of $L$ (\ie, the number of intermediate features). (b) Visual results generated from different end layers. We define $L=0$ as a special case that our method degenerates into merely optimizing the latent vectors $\mathbf{w}$.}
\label{layer}
\end{figure}

\subsection{The Number of Optimized Layers}

To obtain the highest attack performance, the number of intermediate features $L$ should be explored before conducting the major experiments. When $L$ takes a small value, there is a risk of underfitting as we merely optimize the intermediate features of the previous few layers to reconstruct the target images, especially in the OOD scenario. 
In contrast, when $L$ is too large, the latter layers have a greater influence on the local details\cite{karras2019style}, which may lead to overfitting to the target model in some details and produce unrealistic images.
Therefore, we must balance underfitting and overfitting when choosing $L$. We conduct a simple attack on only $10$ classes for each combination of public and private datasets to select $L$ according to the results.
For instance, Figure \ref{layer}(a) shows the $Acc@1$ result for GAN prior pre-trained on FFHQ against the target DenseNet-169 trained on CelebA. The $Acc@1$ reaches the highest when $L=3$. Hence, we keep this configuration in conducting the following experiments.

\begin{table}[!ht]
    \setlength{\tabcolsep}{5pt}
    \normalsize
    \centering
    \caption{Comparison of our method with state-of-the-art methods against ResNet-18 trained on FaceScrub. }
    \label{table:main-facescrub-resnet18}
    \begin{threeparttable} 
    \resizebox{\linewidth}{!}{
    \begin{tabular}{ccccccc}
        \toprule
         \textbf{Public Dataset} & \textbf{Method} & $\uparrow{Acc@1}$ &  $\uparrow{Acc@5}$  & $\downarrow\mathbf{\delta}_{face}$ & $\downarrow\mathbf{\delta}_{eval}$ & $\downarrow${FID} \\ \midrule
         \multirow{6}{*}{FFHQ} 
         & GMI\cite{zhang2020secret} & $0.131$  & $0.339$  & $1.260$  & $149.530$  & $77.800$  \\ 
        ~ & KEDMI\cite{chen2021knowledge} &  $0.127$  & $0.317$  & $1.155$  & $186.409$  & $144.195$  \\ 
        % ~ & VMI\cite{wang2021variational} & $0.616$  & $0.726$  & $0.950$  & $147.480$  & $63.270$  \\ 
        ~ & PPA\cite{struppek2022plug} & $0.962$ &	$\textbf{0.996}$ &	$0.707$&	$117.834$&	$41.688$  \\ 
        ~ & LOMMA+GMI\cite{nguyen2023re} & $0.828$  & $0.945$  & $0.784$  & $126.178$  & $55.840$ \\
        ~ & LOMMA+KEDMI\cite{nguyen2023re} & $0.549$  & $0.814$  & $0.916$  & $217.991$  & $114.045$\\
        ~ & PLGMI\cite{yuan2023pseudo} & $0.758$  & $0.928$  & $0.676$  & $214.978$  & $154.497$  \\ 
        ~ & \textbf{IF-GMI(ours)} & $\textbf{0.979}$  & $\textbf{0.996}$  & $\textbf{0.667}$  & $\textbf{112.915}$  & $\textbf{40.581}$  \\ \midrule 
        \multirow{6}{*}{MetFaces} 
        & GMI\cite{zhang2020secret} & $0.038$  & $0.136$  & $1.361$  & $161.036$  & $114.648$  \\
        ~ & KEDMI\cite{chen2021knowledge} & $0.003$  & $0.017$  & $1.651$  & $212.952$  & $347.468$  \\ 
        % ~ & VMI\cite{wang2021variational} & ~ & ~ & ~ & ~ & ~ \\ 
        ~ & PPA\cite{struppek2022plug} & $0.628$  & $0.854$  & $1.035$  & $146.749$  & $\textbf{62.518}$  \\ 
        ~ & LOMMA+GMI\cite{nguyen2023re} & $0.160$  & $0.361$  & $1.220$  & $156.297$  & $101.600$ \\
        ~ & LOMMA+KEDMI\cite{nguyen2023re} &  $0.002$  & $0.020$  & $1.623$  & $214.883$  & $333.572$ \\
        ~ & PLGMI\cite{yuan2023pseudo} & $0.438$  & $0.731$  & $\textbf{0.796}$  & $205.222$  & $245.208$  \\ 
        ~ & \textbf{IF-GMI(ours)} & $\textbf{0.949}$  & $\textbf{0.992}$  & $0.838$  & $\textbf{120.354}$  & $68.107$ \\
        \bottomrule
    \end{tabular}
    }
    \end{threeparttable}
    % \vspace{-1em}
\end{table}

\subsection{Comparison with Previous State-of-the-art Attacks}

We compare our method with state-of-the-art MI attack methods, including GMI\cite{zhang2020secret}, KEDMI\cite{chen2021knowledge}, PPA\cite{struppek2022plug}, LOMMA\cite{nguyen2023re} and PLGMI\cite{yuan2023pseudo}. Note that LOMMA \cite{nguyen2023re} is a plug-and-play technique designed to augment existing attack methods. We use their original setup where LOMMA is integrated with GMI and KEDMI as our baselines.

The GAN structures employed by GMI, KEDMI, and PLGMI are inherently limited to generating images at a resolution of $64 \times 64$ pixels. To ensure a fair comparison, we adopt the same operation used in PPA \cite{struppek2022plug}, which modifies the architecture of the generators and discriminators to enable the generation of images at an enhanced resolution of $256 \times 256$ pixels, \ie, adding two extra upsampling layers for the generator and two downsampling layers for the discriminator respectively.

\begin{figure}[tbp]
\centerline{\includegraphics[width= 1.2\columnwidth]{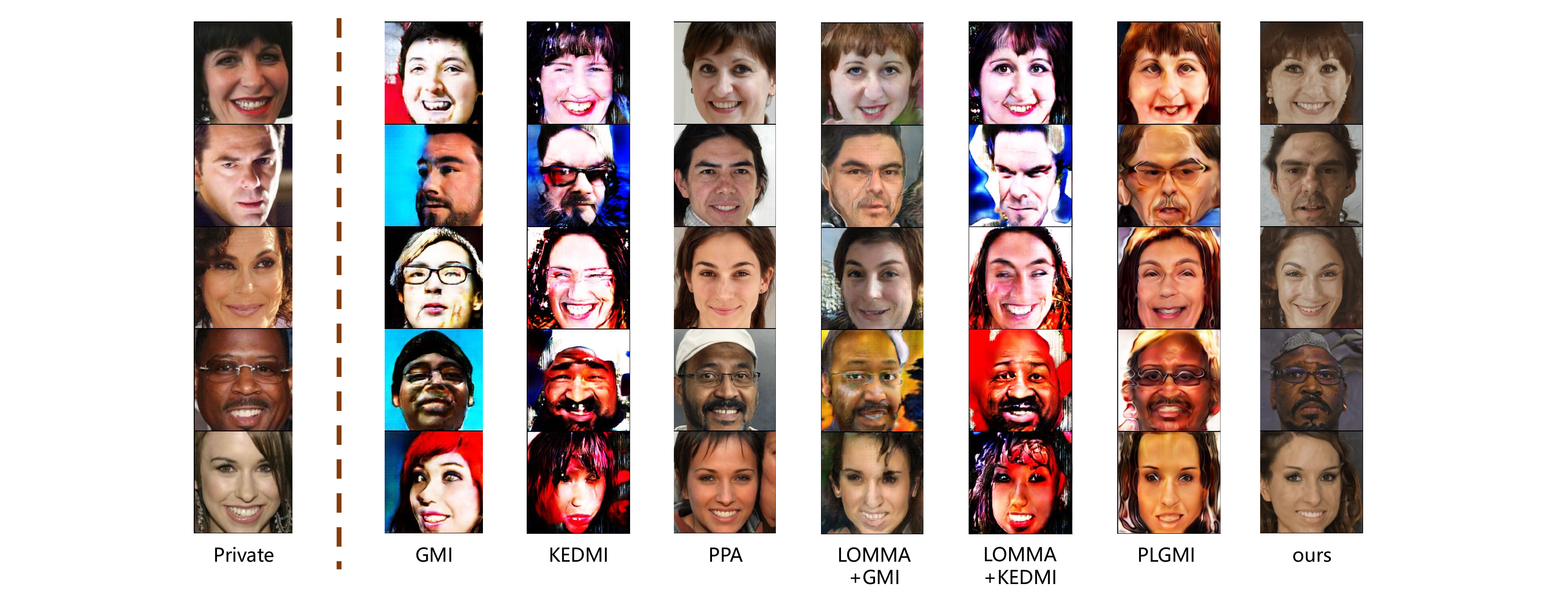}}
\caption{Visual comparison of reconstructed images from different methods against the ResNet-18 trained on FaceScrub. The first column shows ground truth images of the target class in the private dataset.}
\label{visual}
\end{figure}

We provide quantitative results against ResNet-18\cite{resnet} trained on the FaceScrub dataset in Table \ref{table:main-facescrub-resnet18}. 
% The generators are trained on FFHQ and MetFaces datasets respectively. 
We can observe that our method achieves significant improvements over previous methods. Especially when the generator is trained on MetFaces, IF-GMI remarkably improves the $Acc@1$ by $15.1\%$ and the $Acc@5$ is nearly to 100\%. Moreover, our method generally achieves a lower feature distance than baselines between reconstructed samples and private data. For instance, we reduce the distance by more than $10\%$ compared to the PPA on the MetFaces dataset. 
Notably, the MetFaces dataset is composed of artworks and thus has a larger distributional shift with real human faces compared with the FFHQ dataset. We note that this severely reduces the reconstruction performance of previous attack methods, while our proposed method still exhibits outstanding performance, highlighting the excellent generalization ability of our approach.
Visualization results of the recovered images using generators trained on FFHQ are shown in Figure \ref{visual}. Compared with previous methods, our reconstructed images have higher fidelity and realism, demonstrating the superiority of exploiting GAN's intermediate features.

\subsection{Comparison under different target datasets and models} 

To validate the effectiveness of the proposed method, we conducted extensive experiments on various datasets using different target models with different architectures. We chose the PPA method as our baseline for comparison due to its comprehensive performance in both accuracy and fidelity. Additional experimental results are in the Appendix \ref{app:combination}.

\begin{table}[!ht]
    \setlength{\tabcolsep}{5pt}
    \normalsize
    \centering
    \caption{Comparison results against ResNet-152 trained on CelebA. }
    \label{table:main-celeba-resnet152}
    \begin{threeparttable} 
    \resizebox{\linewidth}{!}{
    \begin{tabular}{ccccccc}
        \toprule
         \textbf{Public Dataset} & \textbf{Method} & $\uparrow{Acc@1}$ &  $\uparrow{Acc@5}$  & $\downarrow\mathbf{\delta}_{face}$ & $\downarrow\mathbf{\delta}_{eval}$ & $\downarrow${FID} \\ \midrule
         \multirow{2}{*}{FFHQ} 
        ~ & PPA & $0.806$ &	$0.946$ & $0.736$&	$\textbf{312.580}$&	$40.430$  \\  
        ~ & \textbf{IF-GMI(ours)} & $\textbf{0.912}$  & $\textbf{0.982}$  & $\textbf{0.678}$  & $314.392$  & $\textbf{30.685}$  \\ \midrule 
        \multirow{2}{*}{MetFaces} 
        ~ & PPA & $0.396$  & $0.643$  & $1.063$  & $387.810$  & $\textbf{74.030}$  \\ 
        ~ & \textbf{IF-GMI(ours)} & $\textbf{0.784}$  & $\textbf{0.929}$  & $\textbf{0.835}$  & $\textbf{340.894}$  & $74.504$ \\
        \bottomrule
    \end{tabular}
    }
    \end{threeparttable}
    % \vspace{-1em}
\end{table}
\begin{table}[!ht]
    \setlength{\tabcolsep}{5pt}
    \normalsize
    \centering
    \caption{Comparison results against different target models trained on FaceScrub with the public dataset being MetFaces.}
    \label{table:main-facesrub-more_targets}
    \begin{threeparttable} 
    \resizebox{\linewidth}{!}{
    \begin{tabular}{ccccccc}
        \toprule
         \textbf{Target Model} & \textbf{Method} & $\uparrow{Acc@1}$ &  $\uparrow{Acc@5}$  & $\downarrow\mathbf{\delta}_{face}$ & $\downarrow\mathbf{\delta}_{eval}$ & $\downarrow$FID \\ \midrule
         \multirow{2}{*}{ResNet-152} 
         & PPA & 0.731 & 0.920 & 0.966 & 139.380 & \textbf{68.540} \\ 
         & \textbf{IF-GMI(ours)} & \textbf{0.904} & \textbf{0.984} & \textbf{0.882} & \textbf{138.752} & 69.937  \\
         \midrule 
         \multirow{2}{*}{ResNeSt-101}
         & PPA & 0.750 & 0.927 & 0.979 & 137.170 & 88.660 \\ 
         & \textbf{IF-GMI(ours)} & \textbf{0.922} & \textbf{0.983} & \textbf{0.884} & \textbf{132.609} & \textbf{76.195}  \\
         \midrule 
         \multirow{2}{*}{DenseNet-169}
         & PPA & 0.798 & 0.948 & 0.938 & 129.440 & \textbf{77.520} \\ 
         & \textbf{IF-GMI(ours)} & \textbf{0.933} & \textbf{0.987} & \textbf{0.851} & \textbf{125.050} & 82.123  \\
         \bottomrule
    \end{tabular}
    }
    \end{threeparttable}
    % \vspace{-1em}
\end{table}

As shown in Table \ref{table:main-celeba-resnet152}, our proposed IF-GMI maintains superiority in most metrics against the ResNet-152 trained on the CelebA. Our method achieves a remarkable increase of $10.6\%$ in $Acc@1$ and significantly reduces the FID value using the StyleGAN2 trained on FFHQ. When utilizing the MetFaces StyleGAN2, our method still achieves much better results than the baseline despite a larger distributional shift, including a $38.8\%$ increase in $Acc@1$ and competitive feature distance.
In addition to ResNet-18, we evaluate the performance of the proposed method on more target models trained on FaceScrub, including ResNet-152, ResNeSt-101, and DenseNet-169. Benefiting from the fully utilized generative prior, our method achieves $13\% \sim 17\%$ improvement in $Acc@1$ metrics than the baselines and also achieves better results in most of the other metrics, as illustrated in Table \ref{table:main-facesrub-more_targets}.

% Besides the ResNet-18, we select more target models trained on FaceScrub to evaluate the performance of the proposed method, including ResNet-152, ResNeSt-101, and DenseNet-169. With MetFaces StyleGAN2 as prior, our method keeps the metric $Acc@1$ $13\% \sim 17\%$ higher and outperforms the baseline in most metrics, as shown in Table \ref{table:main-facesrub-more_targets}.

The results presented above demonstrate that our method maintains outstanding attack performance in a variety of settings, exhibiting excellent generalizability and transferability. We also provide additional experimental results on more datasets and architectures in the Appendix \ref{app:combination}.

\begin{table}[!ht]
    \setlength{\tabcolsep}{5pt}
    \normalsize
    \centering
    \caption{Ablation study performed on ResNet-152 trained on CelebA dataset with FFHQ as the public dataset. IF-GMI-\textit{i} removes the intermediate feature optimization and only searches the latent space. IF-GMI-\textit{l} removes the $l_1$ ball constraint compared to IF-GMI.}
    \label{table:ablation-ffhq-celeba-densenet169}
    \begin{threeparttable} 
    \resizebox{.8\linewidth}{!}{
    \begin{tabular}{cccccc}
        \toprule
         \textbf{Method} & $\uparrow{Acc@1}$ &  $\uparrow{Acc@5}$  & $\downarrow\mathbf{\delta}_{face}$ & $\downarrow\mathbf{\delta}_{eval}$ & $\downarrow${FID} \\ \midrule
         % resnet152 - celeba - ffhq
         IF-GMI-\textit{i} & 0.803 & 0.928 & 0.732 & \textbf{314.275} & 43.576 \\ 
         IF-GMI-\textit{l} & 0.945 & 0.992 & 0.678 & 315.278 & 37.528 \\ 
        IF-GMI & \textbf{0.947} & \textbf{0.993} & \textbf{0.677} & 315.032 & \textbf{37.461} \\ 
         % densenet - facescrub - ffhq
         % IF-GMI-\textit{i} & 0.716 & 0.894 & 0.759 & \textbf{316.297} & 46.447 \\ 
         % IF-GMI-\textit{l} & 0.839 & 0.964 & \textbf{0.724} & 318.050 & \textbf{42.852} \\
         % IF-GMI & \textbf{0.846} & \textbf{0.965} & \textbf{0.724} & 317.667 & 42.970 \\
         % densenet - facescrub - metfaces
        % IF-GMI-\textit{i} & 0.578 & 0.817 & 1.067 & 147.009 & \textbf{83.801} \\
        % IF-GMI-\textit{l} & 0.954 & 0.993 & 0.807 & 119.121 & 107.543 \\
        % IF-GMI & \textbf{0.957} & \textbf{0.994} & \textbf{0.806} & \textbf{118.815} & 106.873 \\
        \bottomrule
    \end{tabular}
    }
    \end{threeparttable}
    % \vspace{-1em}
\end{table}

\subsection{Ablation Studies}
\label{main:ablation}
To estimate the contributions from each component in our method, we conduct ablation studies on the ResNet-152 trained on the CelebA dataset using the StyleGAN2 trained on FFHQ. The results are presented in Table \ref{table:ablation-ffhq-celeba-densenet169}. More ablation studies are listed in the Appendix \ref{app:ablation}.

\textbf{Intermediate Features Optimization.} We merely remove the intermediate features optimization from our pipeline while keeping the remaining parameters unchanged. As shown in the first row of Table \ref{table:ablation-ffhq-celeba-densenet169}, it leads to degradation up to $14\%$ in $Acc@1$ and much worse FID without this technique, demonstrating the superiority of utilizing the hierarchical features of intermediate layers.

\textbf{$l_1$ Ball Constraint.}To avoid unreal image generation, we introduce the $l_1$ ball constraint into the intermediate features optimization. By observing the results shown in the second row of Table \ref{table:ablation-ffhq-celeba-densenet169}, the $l_1$ ball is beneficial in improving the performance in all metrics. Thus, we demonstrate the necessity of restricting the intermediate features within the $l_1$ ball constraint.

\section{Conclusion}

We proposed IF-GMI, a novel model inversion attack that performs effective attack in the OOD scenario. Surpassing the limitation of treating the generator as a black-box, we studied the structure and decomposed the generator into hierarchical layers, extending the optimization space from latent code to intermediate features to generate stable and high-quality images. Moreover, to avoid generating low-fidelity images, we applied a ${l}_1$ ball constraint to the optimization process. Through our extensive experiments, we demonstrated that the proposed IF-GMI achieves the state-of-the-art attack accuracy while generating samples with high fidelity and diversity. 

Our exploration of enhanced utilization of intermediate features in the GAN prior contributes to advances in MI attack field, paving the way to more practical employment for MI attacks. We hope this paper can raise concerns about privacy leakage risk of released pre-trained models and facilitate more response to the threat of MI attacks. 

\section*{Acknowledgments}
This work is supported in part by the National Natural Science Foundation of China under grant 62171248, 62301189, Guangdong Basic and Applied Basic Research Foundation under grant 2021A1515110066, the PCNL KEY project (PCL2021A07), and Shenzhen Science and Technology Program under Grant JCYJ20220818101012025, 
RCBS20221008093124061, GXWD20220811172936001.

% \section{Camera-Ready Manuscript Preparation}
% \label{sec:manuscript}
% This information will follow after paper decisions have been announced.

% \section{Conclusion}
% The paper ends with a conclusion. 

% \clearpage\mbox{}Page \thepage\ of the manuscript.
% \clearpage\mbox{}Page \thepage\ of the manuscript.
% \clearpage\mbox{}Page \thepage\ of the manuscript.
% \clearpage\mbox{}Page \thepage\ of the manuscript.
% \clearpage\mbox{}Page \thepage\ of the manuscript. This is the last page.
% \par\vfill\par
% Now we have reached the maximum length of an ECCV \ECCVyear{} submission (excluding references).
% References should start immediately after the main text, but can continue past p.\ 14 if needed.
% TODO REVIEW/FINAL: This \clearpage needs to be removed from both review and camera-ready versions.

% ---- Bibliography ----
%
% BibTeX users should specify bibliography style 'splncs04'.
% References will then be sorted and formatted in the correct style.
%
\bibliographystyle{splncs04}
% \bibliography{main}

\appendix
\setcounter{table}{0}   %从0开始编号，显示出来表会A1开始编号
\setcounter{figure}{0}
\renewcommand{\thetable}{A\arabic{table}}
\renewcommand{\thefigure}{A\arabic{figure}}
\section{Additional Experimental Results}
In this section, we report extra results for our experiment, which are not shown in the main part of the paper.
\subsection{More Comparison with PLGMI \cite{yuan2023pseudo}}
While the reconstructed images from PLGMI have low fidelity and high FID \cite{fid} metric, they achieve relatively high accuracy. To make a comprehensive comparison, we evaluate the attack performance of our method against the state-of-the-art PLGMI under more experimental settings. 
Table \ref{table:plg-metfaces-facescrub} states the results on different target models, including ResNet-152 \cite{resnet}, ResNeSt-101 \cite{zhang2022resnest}, and DenseNet-169 \cite{densenet}. Our method maintains superiority in all metrics compared to PLGMI under the aforementioned settings, aligning with the conclusion that our method achieves new state-of-the-art attack performance.
\begin{table}[!ht]
    \setlength{\tabcolsep}{5pt}
    \normalsize
    \centering
    \caption{Comparison results with PLGMI against different target models trained on FaceScrub \cite{facescrub} with the public dataset being MetFaces \cite{karras2020training}.}
    \label{table:plg-metfaces-facescrub}
    \begin{threeparttable} 
    \resizebox{.8\linewidth}{!}{
    \begin{tabular}{ccccccc}
        \toprule
         \textbf{Target Model} & \textbf{Method} & $\uparrow{Acc@1}$ &  $\uparrow{Acc@5}$  & $\downarrow\mathbf{\delta}_{eval}$ & $\downarrow\mathbf{\delta}_{face}$ & $\downarrow$FID \\ \midrule
         \multirow{2}{*}{ResNet-152} 
         & PLGMI & $0.429$ & $0.708$ & $0.805$ & $243.868$ & $166.614$ \\ 
         & \textbf{Ours} & \textbf{0.904} & \textbf{0.984} & \textbf{0.882} & \textbf{138.752} & \textbf{69.937}  \\
         \midrule 
         \multirow{2}{*}{ResNeSt-101}
         % 0.847	0.968	1.645	340.903	124.985
         & PLGMI & $0.571$ & $0.791$ & $0.814$ & $234.669$ & $213.955$ \\ 
         & \textbf{Ours} & \textbf{0.922} & \textbf{0.983} & \textbf{0.884} & \textbf{132.609} & \textbf{76.195}  \\
         \midrule 
         \multirow{2}{*}{DenseNet-169}
         & PLGMI & 0.390 & 0.645 & 0.818 & 198.095 & 222.563\\ 
         & \textbf{Ours} & \textbf{0.933} & \textbf{0.987} & \textbf{0.851} & \textbf{125.050} & \textbf{82.123}  \\
         \bottomrule
    \end{tabular}
    }
    \end{threeparttable}
    % \vspace{-1em}
\end{table}
\begin{table}[h]
    \setlength{\tabcolsep}{5pt}
    \normalsize
    \centering
    \caption{Comparison results against ResNet-152 trained on Stanford Dogs. The GAN prior is pre-trained on AFHQ \cite{afhq} dataset.}
    \label{table:dogs}
    \begin{threeparttable} 
    \resizebox{.6\linewidth}{!}{
    \begin{tabular}{ccccccc}
        \toprule
         \textbf{Method} & $\uparrow{Acc@1}$ &  $\uparrow{Acc@5}$  & $\downarrow\mathbf{\delta}_{face}$ & $\downarrow${FID} \\ \midrule
         % 0.9498	0.9957		59.25	32.04
         PPA & 0.950 &	0.996 &	\textbf{59.250} &	32.040  \\  
         % 0.987833	0.999833		79.6362381	28.9791
         \textbf{Ours} & $\textbf{0.982}$  & $\textbf{1.000}$ & $70.1956$  & $\textbf{27.282}$  \\ 
        \bottomrule
    \end{tabular}
    }
    \end{threeparttable}
    % \vspace{-1em}
\end{table}

\subsection{More Results on Different Combinations of Datasets and Models}
For the facial image classification task, We further conduct more extensive experiments under multiple combinations of public and private datasets for overall evaluation. The PPA \cite{struppek2022plug} and PLGMI \cite{yuan2023pseudo} are selected as the baseline for comparison due to its comprehensive performance in both accuracy and fidelity. The results shown in Table \ref{table:combination} demonstrate the outstanding performance of our method under various scenarios and verify the excellent generalizability for the proposed method.
\label{app:combination}

For the dog breed classification task, Table \ref{table:dogs} compares attack performance against the ResNet-152 \cite{resnet} trained on Stanford Dogs \cite{standforddog}. The encouraging results exhibit the superiority of our method in another task. 

\begin{table}[!ht]
    \setlength{\tabcolsep}{5pt}
    \normalsize
    \centering
    \caption{Comparison of our method with state-of-the-art methods against ResNet-18 trained on FaceScrub in Improved precision and recall, density, and coverage metrics.}
    \label{table:prdc}
    \begin{threeparttable} 
    \resizebox{\linewidth}{!}{
    \begin{tabular}{ccccccc}
        % \toprule
        %  \textbf{Public Dataset} & \textbf{Method} & $\uparrow{Precision}$ &  $\uparrow{Recall}$  & $\uparrow{Density}$ & $\uparrow{Coverage}$\\ \midrule
        % \multirow{6}{*}{FFHQ} 
        %  & GMI\cite{zhang2020secret} & ~ & ~ & ~ & ~ \\ 
        % ~ & KEDMI\cite{chen2021knowledge} & ~ & ~ & ~ & ~ \\ 
        % ~ & PPA\cite{struppek2022plug} & ~ & ~ & ~ & ~  \\ 
        % ~ & LOMMA+GMI\cite{nguyen2023re} & ~ & ~ & ~ & ~  \\
        % ~ & LOMMA+KEDMI\cite{nguyen2023re} & ~ & ~ & ~ & ~  \\
        % ~ & PLGMI\cite{yuan2023pseudo} & ~ & ~ & ~ & ~ \\ 
        % ~ & \textbf{Ours} & ~ & ~ & ~ & ~ \\ \midrule 
        % \multirow{6}{*}{MetFaces} 
        % & GMI\cite{zhang2020secret} & ~ & ~ & ~ & ~  \\
        % ~ & KEDMI\cite{chen2021knowledge} & ~ & ~ & ~ & ~ \\ 
        % ~ & PPA\cite{struppek2022plug} & ~ & ~ & ~ & ~ \\ 
        % ~ & LOMMA+GMI\cite{nguyen2023re} & ~ & ~ & ~ & ~ \\
        % ~ & LOMMA+KEDMI\cite{nguyen2023re} & ~ & ~ & ~ & ~ \\
        % ~ & PLGMI\cite{yuan2023pseudo} & ~ & ~ & ~ & ~ \\ 
        % ~ & \textbf{Ours} & ~ & ~ & ~ & ~ \\
        % \bottomrule
        \toprule
    \textbf{Public Dataset} & \textbf{Method} & $\uparrow{Precision}$ &  $\uparrow{Recall}$  & $\uparrow{Density}$ & $\uparrow{Coverage}$\\ \midrule
\multirow{3}{*}{FFHQ} 
    & PPA\cite{struppek2022plug} & $0.143$ & $0.003$ & $0.336$ & $0.261$  \\ 
~ & PLGMI\cite{yuan2023pseudo} & $0.005$ & $0.000$ & $0.003$ & $0.004$ \\ 
~ & \textbf{Ours} & $\mathbf{0.186}$ & $\mathbf{0.069}$ & $\mathbf{0.462}$ & $\mathbf {0.314}$ \\ \midrule 
\multirow{3}{*}{MetFaces} 
& PPA\cite{struppek2022plug} & $\mathbf {0.208}$ & $0.000$ & $\mathbf {0.421}$ & $0.182$ \\ 
~ & PLGMI\cite{yuan2023pseudo} & $0.001$ & $0.024$ & $0.000$ & $0.000$ \\ 
~ & \textbf{Ours} & $0.142$ & $\mathbf {0.041}$ & $0.329$ & $\mathbf{0.192}$ \\
\bottomrule
    \end{tabular}
    }
    \end{threeparttable}
    % \vspace{-1em}
\end{table}
\subsection{More Evaluation Metrics}
Following PPA, we further compute the improved Precision-Recall \cite{pr} together with Density-Coverage \cite{dc} on a per-class basis to measure the sample diversity. We utilize the Inception-v3 \cite{inceptionv3} model to calculate the four metrics and make a comprehensive comparison with previous approaches, as shown in Table \ref{table:prdc}. Our method achieve highest scores in most comparisons, indicating superior sample diversity among all the methods.
\label{app:prdc}

\subsection{More Visual Results}
\begin{figure}[tbp]
\centerline{\includegraphics[width= 1.2\columnwidth]{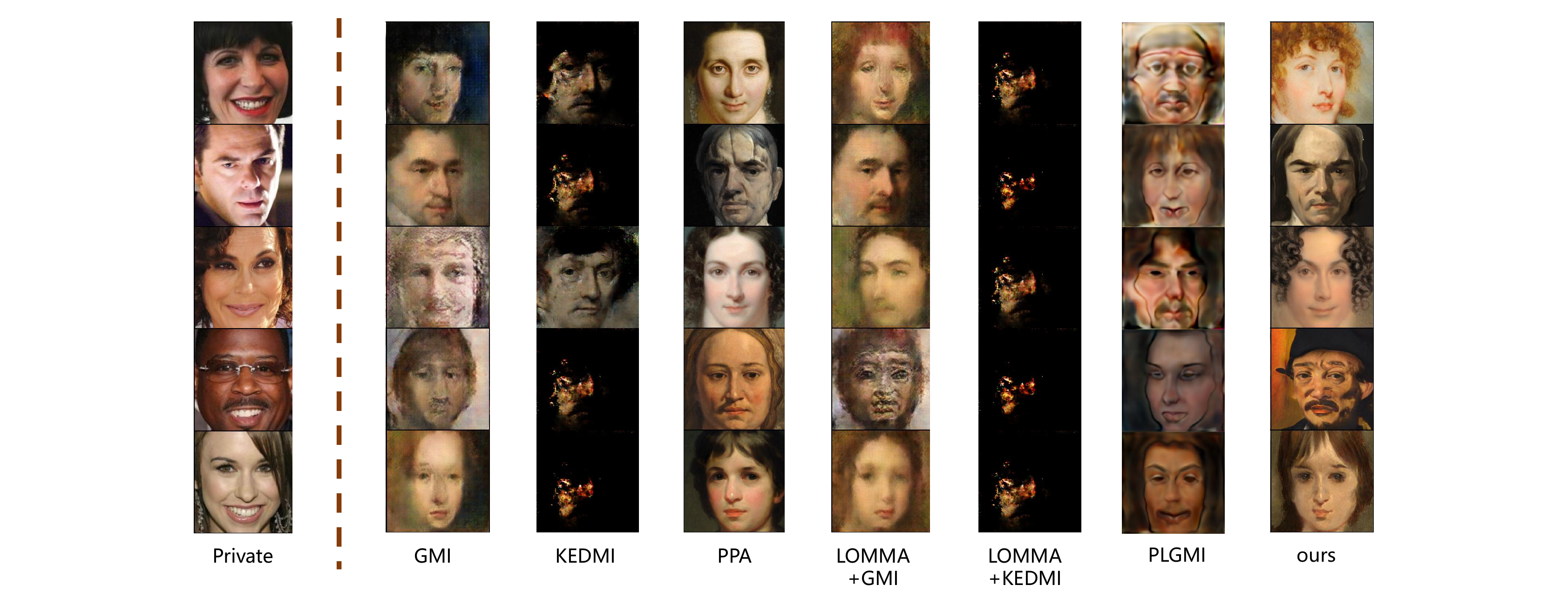}}
\caption{Visual comparison of reconstructed images from different methods against the ResNet-18 \cite{resnet} trained on FaceScrub. The GAN prior is pre-trained on MetFaces. The first column shows ground truth images of the target class in the private dataset.}
\label{visual-metfaces}
\end{figure}

\begin{figure}[tbp]
\centerline{\includegraphics[width= \columnwidth]{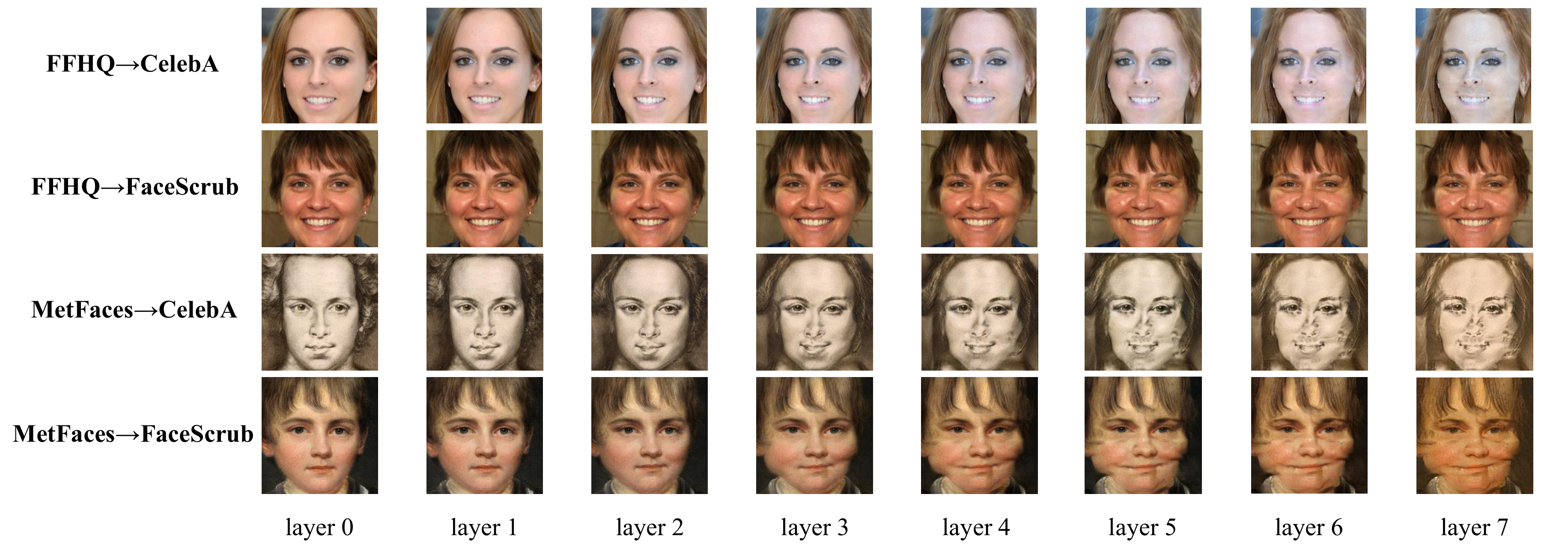}}
\caption{Visual results generated from different end layers. $DatasetA\xrightarrow{}DatasetB$ denotes that the $DatasetA$ is the public dataset and $DatasetB$ is the private dataset.}
\label{layer-2}
\end{figure}

We show more qualitative results generated from the StyleGAN2 \cite{stylegan2} pre-trained on the MetFaces \cite{karras2020training} dataset in Fig \ref{visual-metfaces} and Fig \ref{layer-2}. Fig \ref{visual-metfaces} compares the visual samples from different methods when attacking the ResNet-18 \cite{resnet} trained on FaceScrub. Fig \ref{layer-2} displays the visual images generated from different end layers, showing the gradual change during the intermediate features optimization.

\begin{table}[!htbp]
    \setlength{\tabcolsep}{10pt}
    \centering
    % \vspace{-20pt}
    \caption{Ablation study on $L=1$. The setup aligns with Sec. \ref{main:ablation}.}
    \label{table:ablation_2}
    \begin{threeparttable} 
    \resizebox{.8\linewidth}{!}{
    \begin{tabular}{cccc}
        \toprule
        \textbf{Decomposition} & $\uparrow{Acc@1}$ &  $\uparrow{Acc@5}$ & $\downarrow${FID} \\
        \midrule
        $L=0$ & 0.713 & 0.897 & 46.497 \\ \midrule
        $N=1$ & 0.766 & 0.927 & 44.553 \\
        $N=2$ & 0.795 & 0.940 & \textbf{44.435} \\
        $N=3$ & 0.825 & 0.956 & 45.442 \\
        $N=4$ & \textbf{0.833} & 0.955 & 46.197 \\
        $N=5$ & 0.832 & \textbf{0.957} & 47.628 \\
        $N=6$ & 0.830 & \textbf{0.957} & 50.908 \\
        $N=7$ & 0.829 & 0.956 & 57.009 \\ \midrule
        IF-GMI & \textbf{0.846} & \textbf{0.965} & \textbf{42.970} \\
        \bottomrule
    \end{tabular}
    }
    \end{threeparttable}
    % \vspace{-1em}
\end{table}
\subsection{More Ablation Studies}
To ensure the concrete contributions from each intermediate layer and support the selected number in Sec. 4.2, we conduct further ablation with $L=1$ where $G$ is decomposed in all possible ways. Following StyleGAN2 \cite{stylegan2}, we decompose the generator into style blocks, which serve as the basic intermediate layer units. StyleGAN2 comprises a total of $9$ style blocks. Given that features from later blocks are higher-dimensional and more expensive to optimize, we only select the first seven blocks as intermediate layers. For ablation with $L=1$, we define ``$N=k$'' as the scenario where the blocks before the $(k+1)$-th block are designated as $G_{1}$, and the remaining blocks are designated as $G_{2}$. Therefore, the optimized intermediate feature becomes the output of the $k$-th style block. The settings of $N=k, k\in[1,\dots,7]$ cover all possible decomposition ways when $L=1$. The ablation results are presented in Table \ref{table:ablation_2}. $L=0$ is defined as a special case where only latent vectors are optimized. The last row showcases results under the standard decomposition of $L=3$.

As illustrated in Table \ref{table:ablation_2} in terms of $Acc@1$, $Acc@5$ and FID, the first four intermediate features achieve a good balance between accuracy and image realism. Thus, we opt the first three layers when $L=3$ for improved performance.

\label{app:ablation}
\section{Additional Evaluation on robustness}

To evaluate the robustness of the proposed method, we select the BiDO \cite{peng2022bilateral} as the defense strategy to protect the target model from MI attacks while inducing negligible utility loss. Following the default settings in \cite{peng2022bilateral}, we train a ResNet-152 on FaceScrub as the defensive target model. Without loss of generality, we evaluate our method and baselines on the first $100$ classes of the FaceScrub dataset. See Table \ref{table:defense} for quantitative results. Our method maintains the superior performance and achieves the least decrease of $14.1\%$ in the $Acc@1$ metric after the defense, indicating the distinguished robustness against the BiDO defense.
\section{Additional Experimental Details}
For the target and evaluation models, we use those provided by PPA \cite{struppek2022plug}. The test accuracy of each model is shown in Table \ref{table:acc}.

For initial selection stage, we select $50$ latent vectors from a large batch of candidates. Following PPA, we set the number of candidates as $2000$ for attacking FaceScrub models and $5000$ for attacking CelebA models. For each intermediate feature optimization, we utilize Adam optimizer with a learning rate of 0.005 and $\beta=(0.1,0.1)$ to optimize the intermediate features $\mathbf{f}$ and extended latent vectors $\mathbf{w}$. When attacking the FaceScrub models, we set the optimization steps list as $[50,10,10,10]$ for each intermediate feature. When attacking the CelebA models, the steps list is set as $[70,25,25,25]$. Guided by the theory \cite{daras2021intermediate} that ascended radii of the $l_1$ ball lead to better results, we set the sequence of radius as $[1000, 2000, 3000, 4000]$ for optimization of both $\mathbf{f}$ and extended $\mathbf{w}$.
\begin{table}[htbp]
    \setlength{\tabcolsep}{5pt}
    \normalsize
    \centering
     \caption{Comparison results against more combinations of different datasets and models. Public$\xrightarrow{}$Private denotes that the GAN prior is pre-trained on the public dataset to attack the target model trained on the private dataset.}
    \label{table:combination}
    \begin{threeparttable} 
    \resizebox{\linewidth}{!}{
    \begin{tabular}{cccccccc}
  \toprule
\textbf{Public$\xrightarrow{}$Private} & \textbf{Target Model} & \textbf{Method} & $\uparrow{Acc@1}$ &  $\uparrow{Acc@5}$  & $\downarrow\mathbf{\delta}_{eval}$ & $\downarrow\mathbf{\delta}_{face}$ & $\downarrow${FID} \\
\midrule
\multirow{9}{*}{FFHQ$\xrightarrow{}$FaceScrub} & \multirow{3}{*}{ResNet-152}~ & PPA & $0.927$ & $0.989$ & $0.716$ & $\mathbf {123.250}$ & $46.690$ \\ 
~ & ~ & PLGMI & $0.967$ & $0.999$ & $0.772$ & $170.220$ & $211.217$ \\
~ & ~ & \textbf{Ours} & $\mathbf {0.981}$ & $\mathbf {0.999}$ & $\mathbf {0.691}$ & $131.902$ & $\mathbf {45.703}$ \\ \cmidrule(lr){2-8}
~ & \multirow{3}{*}{ResNeSt-101} & PPA & $0.940$ & $0.992$ & $0.720$ & $\mathbf {119.790}$ & $46.300$ \\ 
~ & ~ & PLGMI & $0.831$ & $0.950$ & $0.873$ & $148.555$ & $177.807$ \\
~ & ~ & \textbf{Ours} & $\mathbf {0.987}$ & $\mathbf {0.998}$ & $\mathbf {0.660}$ & $121.791$ & $\mathbf {42.768}$ \\ \cmidrule(lr){2-8}
~ & \multirow{3}{*}{DenseNet-169} & PPA & $0.953$ & $0.995$ & $0.687$ & $115.200$ & $46.720$ \\ 
~ & ~ & PLGMI & $0.986$ & $0.998$ & $0.659$ & $145.948$ & $134.844$ \\
~ & ~ & \textbf{Ours} & $\mathbf {0.986}$ & $\mathbf {0.998}$ & $\mathbf {0.653}$ & $\mathbf {114.682}$ & $\mathbf {42.896}$ \\ \midrule
\multirow{9}{*}{FFHQ$\xrightarrow{}$CelebA} & \multirow{3}{*}{ResNet-152} & PPA & $0.806$ & $0.946$ & $0.736$ & $\mathbf {312.580}$ & $40.430$ \\ 
~ & ~ & PLGMI & $0.504$ & $0.739$ & $1.630$ & $876.689$ & $70.991$ \\
~ & ~ & \textbf{Ours} & $\mathbf {0.922}$ & $\mathbf {0.985}$ & $\mathbf {0.680}$ & $315.543$ & $\mathbf {30.394}$ \\ \cmidrule(lr){2-8}
~ & \multirow{3}{*}{ResNeSt-101} & PPA & $0.830$ & $0.954$ & $0.751$ & $299.730$ & $44.040$ \\ 
~ & ~ & PLGMI & $0.871$ & $0.968$ & $1.640$ & $709.896$ & $120.983$ \\
~ & ~ & \textbf{Ours} & $\mathbf {0.935}$ & $\mathbf {0.987}$ & $\mathbf {0.705}$ & $\mathbf {298.363}$ & $\mathbf {35.389}$ \\ \cmidrule(lr){2-8}
~ & \multirow{2}{*}{DenseNet-169} & PPA & $0.731$ & $0.905$ & $0.764$ & $\mathbf {312.320}$ & $43.240$ \\ 
~ & ~ & PLGMI & $0.758$ & $0.921$ & $1.622$ & $627.920$ & $115.409$ \\
~ & ~ & \textbf{Ours} & $\mathbf {0.840}$ & $\mathbf {0.955}$ & $\mathbf {0.726}$ & $314.669$ & $\mathbf {36.568}$ \\ \midrule
\multirow{9}{*}{MetFaces$\xrightarrow{}$CelebA} & \multirow{3}{*}{ResNet-152}~ & PPA & $0.396$ & $0.643$ & $1.063$ & $387.810$ & $\mathbf {74.030}$ \\ 
~ & ~ & PLGMI & $0.183$ & $0.391$ & $1.627$ & $682.042$ & $181.555$ \\
~ & ~ & \textbf{Ours} & $\mathbf {0.817}$ & $\mathbf {0.945}$ & $\mathbf {0.815}$ & $\mathbf {334.843}$ & $81.179$ \\ \cmidrule(lr){2-8}
~ & \multirow{3}{*}{ResNeSt-101} & PPA & $0.371$ & $0.629$ & $1.124$ & $387.610$ & $\mathbf {75.070}$ \\ 
~ & ~ & PLGMI & $0.711$ & $0.896$ & $1.617$ & $654.594$ & $175.523$ \\
~ & ~ & \textbf{Ours} & $\mathbf {0.814}$ & $\mathbf {0.942}$ & $\mathbf {0.897}$ & $\mathbf {319.716}$ & $76.831$ \\ \cmidrule(lr){2-8}
~ & \multirow{3}{*}{DenseNet-169} & PPA & $0.309$ & $0.558$ & $1.096$ & $396.810$ & $\mathbf {81.720}$ \\ 
~ & ~ & PLGMI & $0.443$ & $0.704$ & $1.610$ & $615.914$ & $173.141$ \\
~ & ~ & \textbf{Ours} & $\mathbf {0.670}$ & $\mathbf {0.868}$ & $\mathbf {0.893}$ & $\mathbf {341.303}$ & $82.597$ \\
\bottomrule
        % \toprule
        % \textbf{Public$\xrightarrow{}$Private} & \textbf{Target Model} & \textbf{Method} & $\uparrow{Acc@1}$ &  $\uparrow{Acc@5}$  & $\downarrow\mathbf{\delta}_{eval}$ & $\downarrow\mathbf{\delta}_{face}$ & $\downarrow${FID} \\
        % \midrule
        % \multirow{6}{*}{FFHQ$\xrightarrow{}$FaceScrub} & \multirow{2}{*}{ResNet-152}~ & PPA & ~ & ~ & ~ & ~ & ~ \\ 
        % ~ & ~ & \textbf{Ours} & ~ & ~ & ~ & ~ & ~ \\ \cmidrule(lr){2-8}
        % ~ & \multirow{2}{*}{ResNeSt-101} & PPA & ~ & ~ & ~ & ~ & ~ \\ 
        % ~ & ~ & \textbf{Ours} & ~ & ~ & ~ & ~ & ~ \\ \cmidrule(lr){2-8}
        % ~ & \multirow{2}{*}{DenseNet-169} & PPA & ~ & ~ & ~ & ~ & ~ \\ 
        % ~ & ~ & \textbf{Ours} & ~ & ~ & ~ & ~ & ~ \\ \midrule
        % \multirow{6}{*}{FFHQ$\xrightarrow{}$CelebA} & \multirow{2}{*}{ResNet-152} & PPA & ~ & ~ & ~ & ~ & ~ \\ 
        % ~ & ~ & \textbf{Ours} & ~ & ~ & ~ & ~ & ~ \\ \cmidrule(lr){2-8}
        % ~ & \multirow{2}{*}{ResNeSt-101} & PPA & ~ & ~ & ~ & ~ & ~ \\ 
        % ~ & ~ & \textbf{Ours} & ~ & ~ & ~ & ~ & ~ \\ \cmidrule(lr){2-8}
        % ~ & \multirow{2}{*}{DenseNet-169} & PPA & ~ & ~ & ~ & ~ & ~ \\ 
        % ~ & ~ & \textbf{Ours} & ~ & ~ & ~ & ~ & ~ \\ \midrule
        % \multirow{6}{*}{MetFaces$\xrightarrow{}$CelebA} & \multirow{2}{*}{ResNet-152}~ & PPA & ~ & ~ & ~ & ~ & ~ \\ 
        % ~ & ~ & \textbf{Ours} & ~ & ~ & ~ & ~ & ~ \\ \cmidrule(lr){2-8}
        % ~ & \multirow{2}{*}{ResNeSt-101} & PPA & ~ & ~ & ~ & ~ & ~ \\ 
        % ~ & ~ & \textbf{Ours} & ~ & ~ & ~ & ~ & ~ \\ \cmidrule(lr){2-8}
        % ~ & \multirow{2}{*}{DenseNet-169} & PPA & ~ & ~ & ~ & ~ & ~ \\ 
        % ~ & ~ & \textbf{Ours} & ~ & ~ & ~ & ~ & ~ \\
        % \bottomrule
    \end{tabular}
    }
    \end{threeparttable}
\end{table}
\begin{table}[!htbp]
    \setlength{\tabcolsep}{10pt}
    \centering
    % \vspace{-20pt}
    \caption{The test accuracy of target and evaluation models in the experiments. Note that the models and the values in the table are from PPA \cite{struppek2022plug}}
    \label{table:acc}
    \begin{threeparttable} 
    \resizebox{.89\linewidth}{!}{
    \begin{tabular}{cccc}
        \toprule
        \textbf{} & \textbf{FaceScrub} & \textbf{CelebA} & \textbf{Stanford Dogs} \\ \midrule
        \textbf{ResNet-18} & 94.22\% & - & - \\ 
        \textbf{ResNet-152} & 93.74\% & 86.78\% & 71.23\% \\ 
        \textbf{DenseNet-169} & 95.49\% & 85.39\% & 74.39\% \\ 
        \textbf{ResNeSt-101} & 95.35\% & 87.35\% & 75.07\% \\ 
        \textbf{Inception-v3} & 96.20\% & 93.28\% &  79.79\% \\ 
        \bottomrule
    \end{tabular}
    }
    \end{threeparttable}
    % \vspace{-1em}
\end{table}
\begin{table}[!ht]
    \setlength{\tabcolsep}{5pt}
    \normalsize
    \centering
    \caption{Robustness evaluation against the BiDO defense strategy with the ResNet-152 trained on FaceScrub. The GAN prior is pre-trained on MetFaces.}
    \label{table:defense}
    \begin{threeparttable} 
    \resizebox{.8\linewidth}{!}{
    \begin{tabular}{ccccccc}
    \toprule
       \textbf{Method} & $\uparrow{Acc@1}$ &  $\uparrow{Acc@5}$  & $\downarrow\mathbf{\delta}_{eval}$ & $\downarrow\mathbf{\delta}_{face}$ & $\downarrow${FID} \\ \midrule
        PPA & 0.619 & 0.873 & 1.010 & 149.621 & \textbf{69.263} \\ 
        \textbf{Ours} & \textbf{0.906} & \textbf{0.985} & \textbf{0.861} & \textbf{134.634} & 76.108 \\ \midrule
        PPA+BiDO & 0.356 & 0.639 & 1.119 & 167.318 & \textbf{70.759} \\ 
        \textbf{Ours+BiDO} & \textbf{0.765} & \textbf{0.924} & \textbf{0.957} & \textbf{146.949} & 74.197 \\
    \bottomrule
    \end{tabular}
    }
    \end{threeparttable}
\end{table}

\begin{table}[!htbp]
    \setlength{\tabcolsep}{16pt}
    \centering
    % \vspace{-20pt}
    \caption{FID scores between various datasets. Note that the values in the table are from PPA \cite{struppek2022plug}.}
    \label{table:shift}
    \begin{threeparttable} 
    \resizebox{.7\linewidth}{!}{
    \begin{tabular}{ccc}
        \toprule
        \textbf{Dataset 1} & \textbf{Dataset 2} & \textbf{FID} \\
        \midrule
        FFHQ & FaceScrub & 77.90 \\
        FFHQ & CelebA & 59.48 \\
        \midrule
        MetFaces & FaceScrub & 104.33 \\
        MetFaces & CelebA & 93.64 \\
        \bottomrule
    \end{tabular}
    }
    \end{threeparttable}
    % \vspace{-1em}
\end{table} 
% \smallskip

\section{Distributional Shift Between Datasets}
Following PPA \cite{struppek2022plug}, we employ FID scores to measure distributional distance and the corresponding results are summarized in Table \ref{table:shift}. A higher value indicates a greater distributional disparity between datasets, revealing that FFHQ \cite{karras2019style} is an easier OOD scenario compared to the more challenging MetFaces \cite{karras2020training}. Our method is confirmed to be effective across various OOD scenarios, particularly under tougher scenarios.
\section{Limitations \& Future work.} 
Despite the superior performance in most metrics under multiple experimental settings, we find that the generated images achieve relatively high FID scores. The potential reason is that we directly perform backpropagation on the intermediate features that are higher-dimensional than the latent code while utilizing the same loss originally designed for latent vectors. In the future work, we will explore a better solution to design an appropriate optimization strategy for intermediate features. 

\end{document}